\documentclass[10pt]{article} %
\usepackage[preprint]{tmlr}

\usepackage{amsmath,amsfonts,bm}

\def\eqref#1{equation~\ref{#1}}

\def\1{\bm{1}}

\DeclareMathAlphabet{\mathsfit}{\encodingdefault}{\sfdefault}{m}{sl}
\SetMathAlphabet{\mathsfit}{bold}{\encodingdefault}{\sfdefault}{bx}{n}

\usepackage{todonotes}
\usepackage{hyperref}
\usepackage{url}
\usepackage{graphicx}
\usepackage{booktabs}

\title{Towards Reliable Scaling Laws for Tabular Foundation\\ Models}

\newcommand{\asep}{\unskip\quad\ignorespaces}

\author{\name
  Johannes Hog$^{1,*,\dagger}$ \asep
  Neeratyoy Mallik$^{1,2,*}$ \asep
  Steven Adriaensen$^{1}$ \asep
  Jan Hendrik Metzen$^{3}$ \\
  Tom Viering$^{4}$ \asep
  Danny Stoll$^{1}$ \asep
  Frank Hutter$^{3,5,1}$ \\[2pt]
  \addr
  $^1$University of Freiburg \asep
  $^2$Zuse School ELIZA \asep
  $^3$Prior Labs \asep
  $^4$Delft University of Technology \asep \\
  $^5$ELLIS Institute Tübingen 
  $^*$Equal contribution \asep \\
  $^\dagger$Corresponding author: \textup{\texttt{hogj@cs.uni-freiburg.de}}
}

\hypersetup{pdftitle={Towards Reliable Scaling Laws for Tabular Foundation Models},
  pdfauthor={Johannes Hog, Neeratyoy Mallik, Steven Adriaensen, Jan Hendrik Metzen,
             Tom Viering, Danny Stoll, Frank Hutter}}

\def\month{MM}  %
\def\year{YYYY} %
\def\openreview{\url{https://openreview.net/forum?id=XXXX}} %

\begin{document}

\maketitle

\begin{abstract}

Scaling laws are widely used to predict foundation model performance and guide compute allocation. 
However, tabular foundation models (TFMs) differ from conventional foundation models in that they are pretrained on virtually unlimited synthetic data from configurable distributions, while ultimately being evaluated on real-world datasets.
We investigate whether scaling laws for TabPFNv2-style models reliably extrapolate to larger compute budgets and generalize across evaluation settings. 
Our analysis reveals differences between synthetic and real-world scaling behavior and shows that accurately fitting a scaling law does not necessarily yield reliable prescriptions on unseen datasets.
In particular, selecting configurations using too few evaluation datasets leads to systematically over-optimistic performance estimates. 
We further demonstrate that compute-optimal scaling requires expanding the range of synthetic pretraining dataset sizes alongside model size and training compute. 
Increasing the maximum numbers of rows and features sampled during pretraining improves performance and produces more predictable scaling trends. 
These insights guide the training of more efficient TabPFNv2-like models,
using the open-source training pipeline of nanoTabPFN.

\end{abstract}

\section{Introduction}
Tabular data underpins predictive tasks across diverse domains, including healthcare, finance, and scientific discovery~\citep{borisov2022deep,priorlabs2026awesome}.
Unlike conventional models that require dataset-specific training, tabular foundation models (TFMs) achieve state-of-the-art performance on small- to medium-sized datasets without dataset-specific parameter updates~\citep{erickson2026tabarena,purucker2026beyond,jager2026tabpfn}.
Models such as TabPFN~\citep{hollmann-nature25a,jager2026tabpfn}, TabICL~\citep{qu2025tabicl,qu2026tabiclv}, LimiX~\citep{zhang-2025limix}, and Mitra~\citep{zhang2025mitra} learn to perform in-context prediction by pretraining on synthetic datasets sampled from task priors.

The success of TFMs has motivated extensions to time-series forecasting~\citep{hoo2026from}, causal effect estimation~\citep{robertson2025dopfn}, survival prediction~\citep{bohm2026survpfn,qi2026survivalpfn}, and hyperparameter optimization~\citep{muller-icml23a,rakotoarison-icml24a}.
These applications motivate experimentation with task-specific priors and model architectures, increasingly facilitated by open-source implementations such as TabICL~\citep{qu2026tabiclv} and TFM-Playground~\citep{pfefferle2026tfmplayground}.
However, while such experimentation is feasible at small scales, evaluating promising configurations at competitive compute budgets remains expensive, especially as models scale to hundreds of millions of parameters~\citep{kong2026tabfm}.
Scaling laws, widely used to predict performance and guide compute allocation in language modeling~\citep{kaplan-arxiv20a,hoffmann-neurips22a}, offer a framework for bridging this gap.

Previous studies have investigated scaling relationships for TFMs~\citep{zhang-2025limix,ma2025tabdpt,zhang2025mitra}, but accurately fitting observed performance does not guarantee reliable extrapolation to larger compute budgets.
In particular, unconstrained scaling-law formulations can yield predictions outside the valid range of evaluation metrics, such as ROC AUC \citep{zhang-2025limix}.
Selecting compute-optimal configurations on too few evaluation datasets can further yield over-optimistic performance estimates even with good scaling-law fits.
Finally, the synthetic prior design and sampling contain many degrees of freedom that yield vastly different data mixtures while remaining finely controllable. 
This makes the prior itself a scaling axis for the compute-optimal performance of TFMs.
We investigate these challenges using over $6{,}000$ training runs spanning model sizes, compute budgets, synthetic prior configurations, and training hyperparameters.
Our main contributions are:

\begin{itemize}

    \item \textbf{A systematic approach to study tabular scaling.}
    We conduct a comprehensive scaling analysis of TabPFNv2-style models, jointly exploring model scale, training hyperparameters, and synthetic prior configurations. We derive scaling laws and prescriptions for these dimensions, considering both synthetic validation and real-world performance (Section~\ref{sec:reliable-laws}).

    \item \textbf{Compute-optimal scaling of synthetic prior support.}
    To our knowledge, we provide the first systematic investigation 
    where we show that compute-optimal configurations favor increasing the maximum numbers of sampled rows and features
    (Section~\ref{sec:prior-scaling}).

    \item \textbf{Robustness across evaluation settings.}
    We compare scaling trends across synthetic and real-world evaluation data and metrics.
    We further demonstrate that accurate scaling-law fits can conceal substantial evaluation-set selection bias when configurations are selected on too few datasets, and we quantify how this bias decreases as the number of evaluation datasets grows (Section~\ref{sec:transfer}).
\end{itemize}

The scaling-law data and code for reproducing the figures are available on GitHub\footnote{\url{https://github.com/JohannesHog/tfm-scaling-laws}}.

\section{Related Work}

\paragraph{Scaling Laws}
Scaling laws characterize how loss varies with model size, training data, and compute, typically through parametric power-law fits~\citep{kaplan-arxiv20a,hoffmann-neurips22a,sorscher-neurips22a,alabdulmohsin-neurips23a,caballero-iclr23a}.
They are commonly estimated from scaling ladders of smaller training runs to predict performance and compute-optimal configurations at larger scales~\citep{bhagia-colm25a}.
However, scaling-law estimates are sensitive to experimental design, compute accounting, hyperparameter tuning, and fitting choices, potentially undermining extrapolation reliability~\citep{viering2022shape,mohr2022lcdb,porian2024resolving,li2025mis}.
Recent work investigates reliable scaling-law estimation~\citep{choshen2024hitchhiker}, how to select informative small-scale experiments~\citep{lourie-arxiv26a}, and benchmarks for comparing scaling-law methodologies across language and vision-language models~\citep{sermaxhaj2026scan}.
The cost of tuning scaling ladders~\citep{jha-neurips26a} has also motivated modeling how optimal hyperparameters vary with scale~\citep{yang-neurips21a,everett-icml24a,bjorck-iclr25a,ajroldi-arxiv26a}.

\paragraph{Scaling Laws for Tabular Foundation Models}
Recent work has investigated how tabular foundation models scale with model size, pretraining data, and compute. 
TabDPT~\citep{ma2025tabdpt} establishes power-law relationships between model size and real-world pretraining data, while LimiX~\citep{zhang-2025limix} studies scaling on synthetic data. 
Its successor, LimiX-2~\citep{zhang2026limix}, fits downstream performance scaling laws across multiple benchmarks and extrapolates to larger models. 
MegaTabICL~\citep{kaushal2026megatabicl} investigates compute-optimal model sizing and training duration. Beyond these scaling studies, TabFM~\citep{kong2026tabfm} demonstrates strong downstream performance through large-scale synthetic pretraining, while Mitra~\citep{zhang2025mitra} and TabPFN-Wide~\citep{kolberg2025tabpfn} highlight the importance of synthetic prior design. 
While prior work shows consistent scaling trends, it has not sufficiently evaluated predictive reliability beyond the fitted range. 
We assess extrapolation errors and transferability to unseen evaluation datasets, while investigating synthetic prior support as an additional 
scaling dimension.

\section{A Reliable Scaling Analysis for Tabular Foundation Models}
\label{sec:scaling-fits}
We describe our scaling analysis, experimental setup, and fitting procedure, focusing on deriving compute-dependent scaling prescriptions and validating their predictive reliability at unseen scales.
\paragraph{Design of the Scaling Analysis}
We explore a search space spanning eight architectural, training, and synthetic prior parameters, including the maximum number of rows and features sampled during pretraining (see  Table~\ref{tab:hyperparameters}). 
We investigate the trade-off between model size and training data volume, fit scaling laws to the empirical compute-optimal performance frontier, and derive power-law scaling prescriptions for individual parameters. 
To assess their reliability, we apply these prescriptions at unseen compute scales and compare the resulting performance with scaling-law predictions. 
We consider both synthetic validation loss and real-world evaluation metrics, including NLL and ROC AUC.

\paragraph{Tabular Foundation Model Pipeline}
We use TFM-Playground~\citep{pfefferle2026tfmplayground} to train nanoTabPFN~\citep{pfefferle2025nanotabpfn}, a simplified implementation of the TabPFNv2 architecture~\citep{hollmann-nature25a}, on classification tasks sampled from the TabICLv1 MLP prior~\citep{qu2025tabicl}.
We chose this architecture because, at the outset of our study, it was among the few strong TFM architectures with an openly available training pipeline, enabling systematic experimentation with model configurations and synthetic priors.
We evaluate synthetic validation performance on two fixed, held-out sets of small and medium-sized datasets, combining their losses into a single validation score.
We additionally evaluate downstream performance on TabArena~\citep{erickson2026tabarena}, restricting the benchmark to datasets with at most $500$ features and $10{,}000$ samples, consistent with the TabPFNv2 evaluation setting. 
To efficiently obtain scaling measurements at multiple compute budgets, we train with a constant learning rate and save intermediate checkpoints at geometrically spaced FLOP intervals. 
We provide further experimental details in Appendix~\ref{app:experimental_setup}.

\paragraph{Fitting Procedure}
For each evaluation metric, we construct the Pareto frontier over training configurations and intermediate checkpoints to estimate the best observed performance achievable within a given compute budget, excluding configurations dominated by alternatives requiring equal or less compute.
For metrics to be maximized and bounded above by $1$, such as accuracy and ROC AUC, we transform the objective $y$ into $L=1-y$.
We model the resulting loss $L$ as a function of training compute $C$ using,
\begin{equation}
    L(C) = A C^{-\alpha} + L_0,
    \label{eq:scaling}
\end{equation}
where $A>0$ is a scaling coefficient, $\alpha>0$ is the scaling exponent, and $L_0>0$ is the asymptotic loss.
For transformed metrics, this formulation ensures that predictions increase monotonically with compute but never exceed the metric's upper bound
avoiding invalid extrapolations and improving robustness.
Optimization details, initialization settings and the fitting procedure for scaling parameters are provided in Appendix~\ref{app:scaling_fits}.

\section{Results}

\subsection{Reliable Scaling Laws and Prescriptions}
\label{sec:reliable-laws}
\begin{figure}[htbp]
    \centering
    \includegraphics[width=0.75\linewidth]{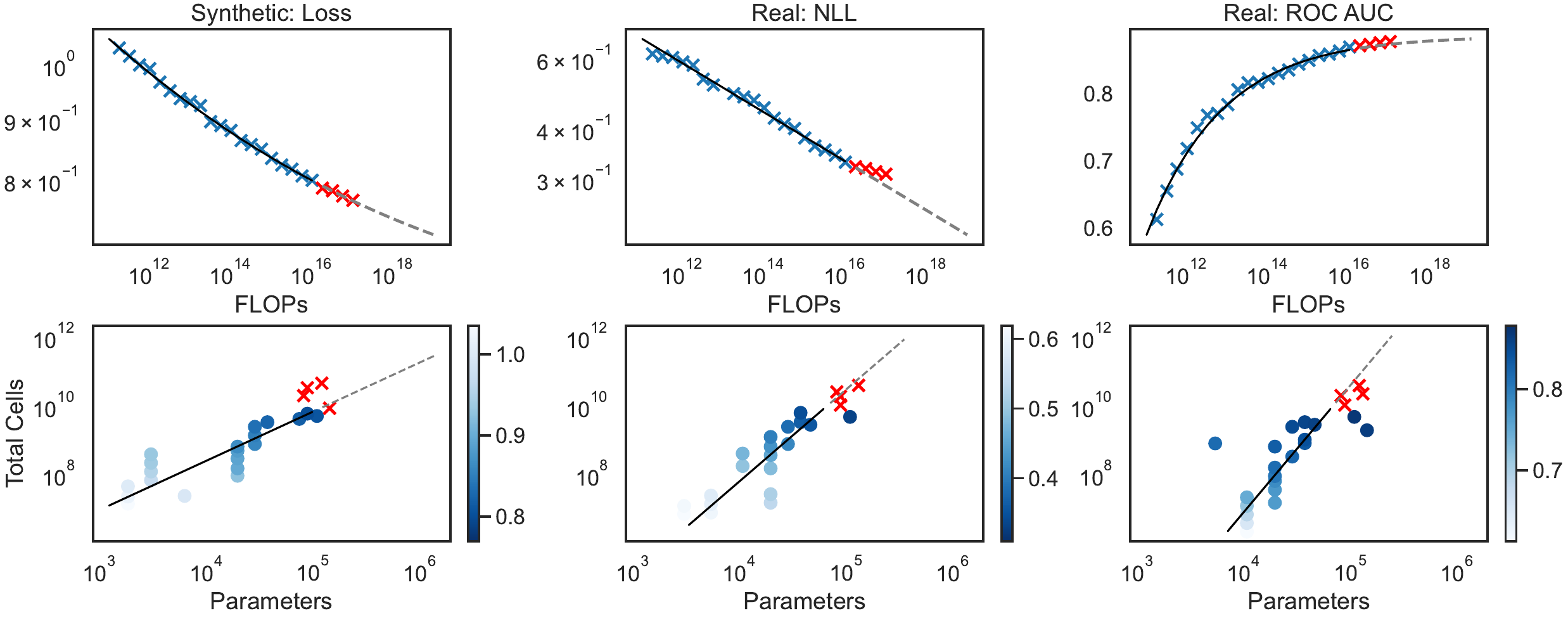}
    \caption{Pareto Trends for Different Metrics. We observe typical compute-optimal performance scaling for both synthetic validation loss and aggregated metrics on real datasets. 
    With more high-compute observations, scaling fits on any performance metric can extrapolate reliably.
    However, the relative scaling trends of data and parameters depend heavily on the choice of metric.
    Solid lines: in-sample fit; Dashed: extrapolation.
    }
    \label{fig:pareto-trends}
\end{figure}

\begin{figure}[htbp]
    \centering
    \begin{tabular}{ccc}
         \includegraphics[width=0.25\columnwidth]{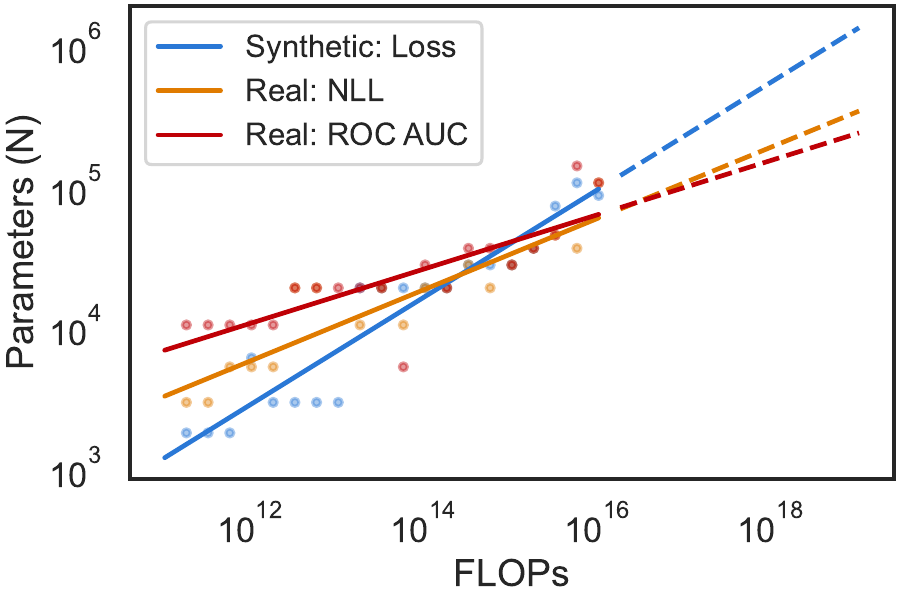} & 
          \includegraphics[width=0.25\columnwidth]{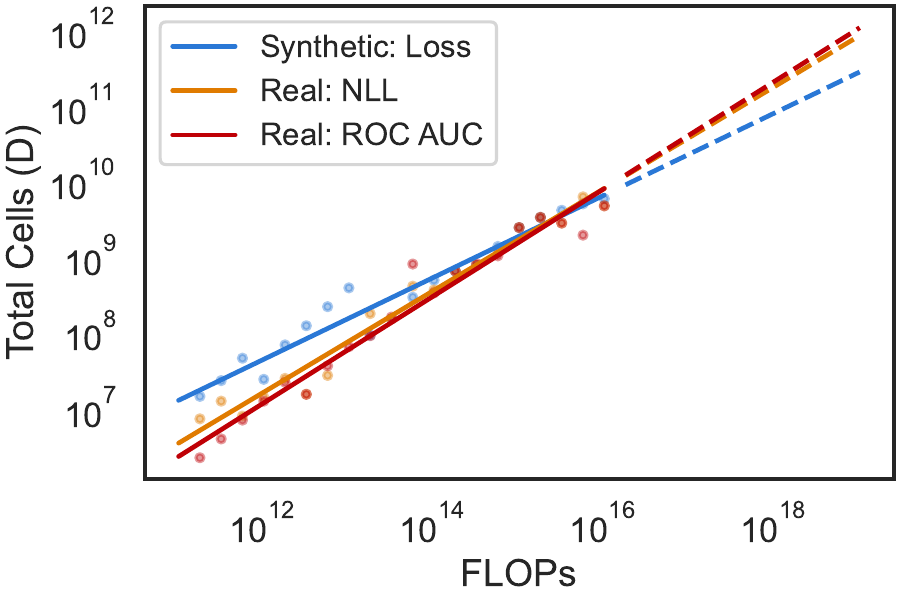} &
        \includegraphics[width=0.25\columnwidth]{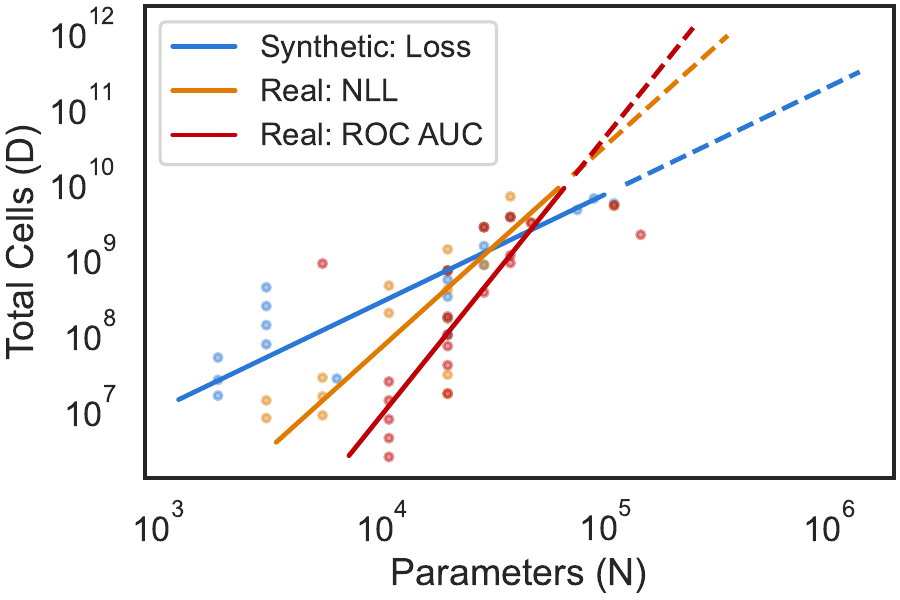}
        
    \end{tabular}  
    \caption{Comparing trends for scaling parameters across synthetic and real-world evaluation. 
    Scaling prescriptions generally agree within real and synthetic data but disagree between synthetic and real-world data.
    Solid lines: in-sample fit; Dashed: extrapolation.
    }
    \label{fig:metric-wise-scaling}
\end{figure}

Compute-optimal scaling performance is the best metric value attained at a given FLOP budget (see FLOP calculation in Appendix~\ref{app:flops}).
In Figure~\ref{fig:pareto-trends} we exhibit the Pareto front of performance over FLOPs for the runs we collect over our scaling grid (Table~\ref{tab:hyperparameters}).
We measure performance over two evaluation sets: a fixed held-out set of synthetic datasets drawn from the prior, and the set of real-world datasets (Section~\ref{sec:scaling-fits}).
We fit the parameters of Eq.~\ref{eq:scaling} on the lower compute scales of our scaling grid (in {\color{blue}blue} in Fig.~\ref{fig:pareto-trends}), and use the higher compute scales of this grid to evaluate the extrapolation quality of this scaling fit (in {\color{red}red} in Fig.~\ref{fig:pareto-trends}; Table~\ref{tab:extrapolation}).
For all considered metrics, we see that performance continues to improve with more compute (top row, Fig.~\ref{fig:pareto-trends}).
From the selected Pareto configurations, we also fit the relationship of compute with the number of parameters, and of compute with the total number of data cells observed.
In the bottom row of Fig.~\ref{fig:pareto-trends}, we show these configurations in the plane of parameters and data cells for each performance metric, and observe that the relative rate of scaling parameters and data depends on the metric.
In Figure~\ref{fig:metric-wise-scaling} we make this comparison explicit by showing the fitted trends of parameters and data cells against compute for each metric (also refer to Figures~\ref{fig:trends_searched}-\ref{fig:trends_derived}).
The fits show that the synthetic evaluation set demands parameter scaling more than the real-world metrics do, whereas performance on real-world benchmarks benefits more from additional synthetic pretraining data.
These results illustrate that the basic signal for scaling can be obtained from any of the considered performance metrics.
However, for a reliable extrapolation prescription, both synthetic and real-world metrics need to be measured.
The synthetic fit and the real-world fit can together give the bracket within which practical large-compute runs can be operated.

\subsection{Scaling the Support of the Training Prior}\label{sec:prior-scaling}

\begin{figure}[htbp]
    \centering
    \begin{tabular}{c}
         \includegraphics[width=0.65\columnwidth]{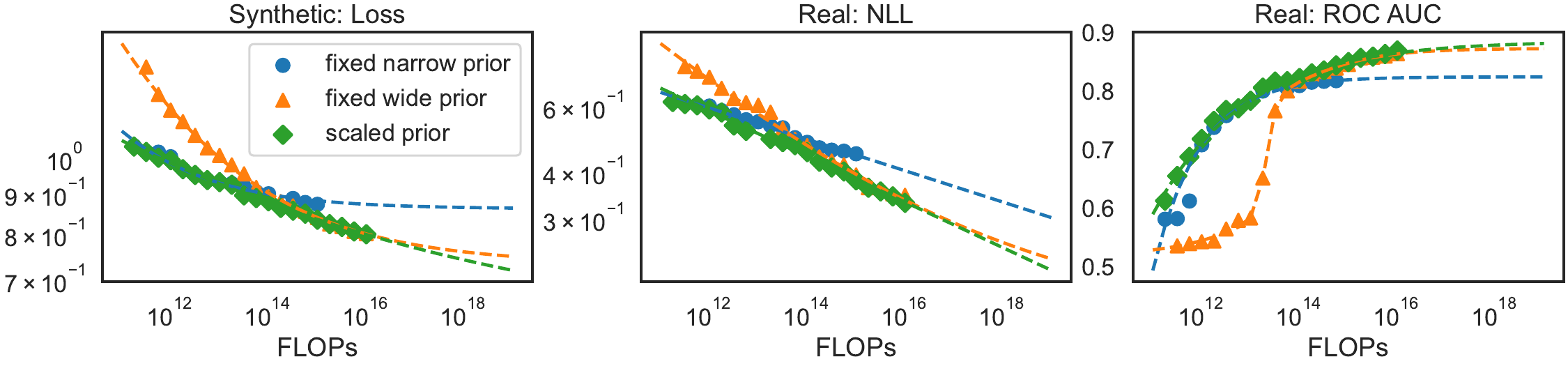} \\
          \includegraphics[width=0.65\columnwidth]{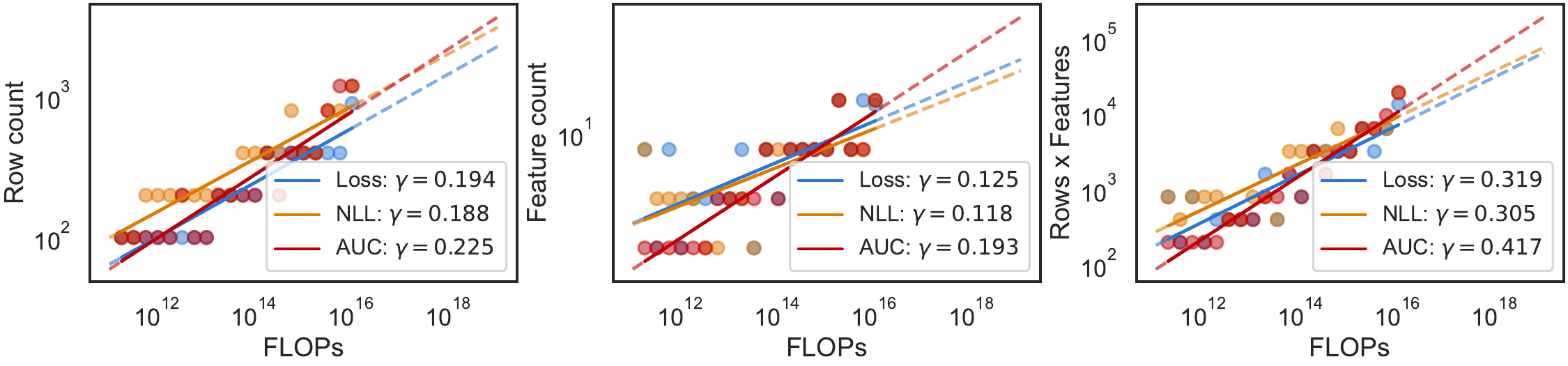}
    \end{tabular}
    \caption{
    The size of the pretraining synthetic dataset has two degrees of freedom: the size of rows and columns. 
    Both are scaling variables that need scaling with compute, irrespective of the metric measured.
    The $\gamma$ is the exponent on the compute when modeling the y-axis metric as a power law $a\,C^{\gamma}$.
    }
    \label{fig:scaling-data-prior}
\end{figure}

The design of the synthetic prior family is a crucial component in the TFM pipeline.
Typically, the prior itself exposes many degrees of freedom that can be configured for performance scaling.
For this experiment, we look only at the role of the size of the synthetic datasets sampled for TFM pretraining, under a fixed prior family.
In Figure~\ref{fig:scaling-data-prior}, we extract Pareto fronts under different data prior conditions from our entire scaling grid collection.
For the \textit{narrow prior}, we restrict the sampling range of dataset sizes (i.e., number of rows and columns) to a small range.
For the \textit{wide prior}, we sample datasets from a wide range (narrow $\subset$ wide).
The \textit{scaled prior} is the \textit{global} Pareto front when no conditioning on dataset size is applied.
All synthetic losses are measured on the same held-out set as in the previous section. 
The key observation is that neither the full wide range nor the narrow restricted range matches the global selection represented by the \textit{scaled prior}. 
The scaled prior follows Eq.~\ref{eq:scaling} across the full compute range, for any metric chosen.
For ROC AUC, the wide prior performance frontier deviates from the typical parametric relationships drastically (see Figure~\ref{fig:big-prior-fits} for more details on the custom logistic fit).
We then take these \textit{scaled prior} Pareto configurations and analyze their dataset properties in the bottom row (Fig.~\ref{fig:scaling-data-prior}), where $\gamma$ denotes the fitted exponent in FLOPs.
Under compute scaling, the Pareto configurations use larger synthetic datasets as compute grows, and that holds for both dimensions of the dataset size: rows and columns.
The relative rate of scaling of these variables depends on the metric measured: synthetic loss and real-world NLL agree closely, whereas the bounded AUC metric prescribes faster growth with compute.

\subsection{Robustness of the Scaling Analysis}
\label{sec:transfer}
\begin{figure}[t]
  \centering
  \includegraphics[width=0.555\linewidth]{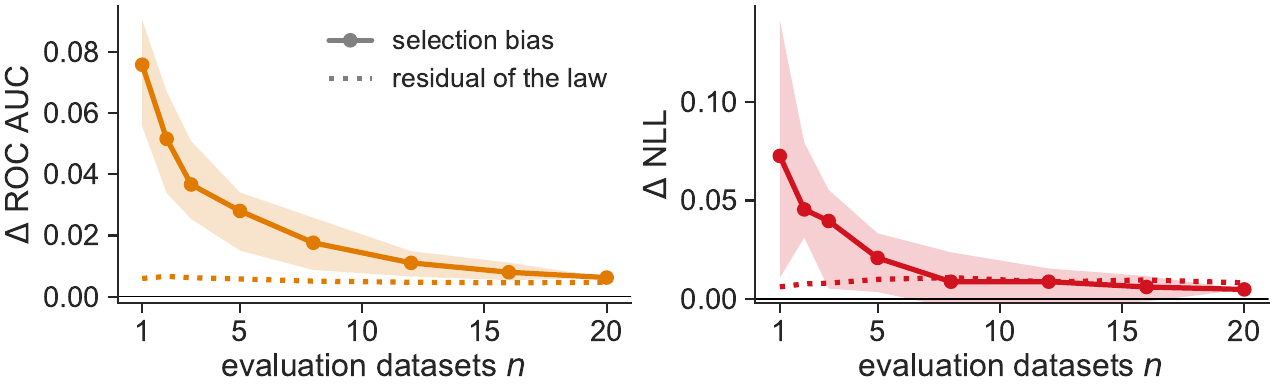}
  \caption{
  Selection bias of the scaling frontier versus the number of evaluation datasets $n$. Solid: gap between the in-sample frontier built from $n$ datasets and the same runs scored on a held-out dataset (median, 10--90\% band over draws). Dotted: in-sample residual of the fitted scaling law, taken as the noise floor. The gap reaches the residual at $n = 8$ for NLL but only at $n = 20$ for ROC AUC.
  }
  \label{fig:eval-datasets}
\end{figure}

Until now, we observed the importance of measuring different evaluation metrics for gauging scaling performance, where we measure the aggregated NLL or ROC AUC on the set of selected datasets.
However, a scaling fit can be overfit when the evaluation set is small.
To investigate this selection effect and validate the robustness of scaling fits, we hold out each evaluation dataset $i$ from the total evaluation set.
We iterate over different subset sizes $n\in\{1, \ldots, 20 \}$, randomly sampling without replacement $n$ datasets, excluding the $i$-th held-out set.
For a given $n$, we obtain the metric scored under the subset of $n$ datasets, and extract the corresponding Pareto frontier.
For these selected Pareto configurations, we measure their score on the single $i$-th held-out set.
Both of these scores are repeatedly computed multiple times for different held-out sets and $n$.
The difference between these two aggregated scores is displayed as the solid line in Figure~\ref{fig:eval-datasets}.
The dotted line measures the in-sample mean residual of the scaling law averaged over the sampled $n$-datasets.
This residual signal is treated as the accepted noise level where we qualify the difference in performance on the sampled-$n$ and the held-out as negligible.
This performance difference falls below the fitting residual at $n=8$ for NLL but only reaches it at $n=20$ for ROC AUC. 
These findings highlight that low fitting error alone does not establish reliable scaling laws and that an appropriately selected held-out evaluation is essential for assessing their generalization and therefore robustness.

\section{Conclusion}\label{sec:conclusion}

In a large-scale scaling grid evaluation of the TabPFNv2 architecture, we studied three choices that shape a scaling analysis: the metric that is measured and modeled, the number of real-world evaluation datasets, and the size of the synthetic datasets sampled from the prior. 
TFMs follow typical scaling trends under every metric we measured, but synthetic and real-world metrics prescribe different allocations of parameters and data, so both must be measured for reliable extrapolation. 
The real-world evaluation set must be large enough to suppress selection bias. 
Most importantly, the size of the synthetic datasets is itself a scaling variable and belongs on the scaling ladder. 
This opens a clear future direction of developing functional forms that include the prior as a variable and give sharper prescriptions in practice. 
Our work offers a blueprint for making TFM scaling practice reliable.

\subsubsection*{Acknowledgments}
\textbf{JH} acknowledges funding by the Deutsche Forschungsgemeinschaft (DFG, German Research Foundation) under SFB 1597 (SmallData), grant number 499552394.
\textbf{NM} is supported by the Konrad Zuse School of Excellence in Learning and Intelligent Systems (ELIZA) through the DAAD programme Konrad Zuse Schools of Excellence in Artificial Intelligence, sponsored by the Federal Ministry of Education and Research.
\textbf{FH} acknowledges the financial support of the Hector Foundation.
We acknowledge funding by the European Union via ERC Consolidator Grant DeepLearning 2.0, grant no.~101045765 and via the Horizon Europe research and innovation programme under grant agreement no.~ 101214398 (ELLIOT).
Funded by the European Union. Views and opinions expressed are however those of the author(s) only and do not necessarily reflect those of the European Union, the European Commission, or the European Research Council. Neither the European Union nor the European Commission nor the European Research Council can be held responsible for them.
This research was funded by the Deutsche Forschungsgemeinschaft (DFG, German Research Foundation) under grant number 539134284, through EFRE (FEIH\_2698644) and the state of Baden-Württemberg.
The authors gratefully acknowledge the computing time made available to them on the high-performance computer at the NHR Center of TU Dresden. This center is jointly supported by the Federal Ministry of Research, Technology and Space of Germany and the state governments participating in the NHR (www.nhr-verein.de/unsere-partner).
We acknowledge the EuroHPC Joint Undertaking for awarding this project access to the EuroHPC supercomputer LUMI, hosted by CSC (Finland) and the LUMI consortium through a EuroHPC Regular Access call.

\begin{center}
\hfill
\begin{minipage}{0.24\textwidth}
\centering
\includegraphics[width=\textwidth]{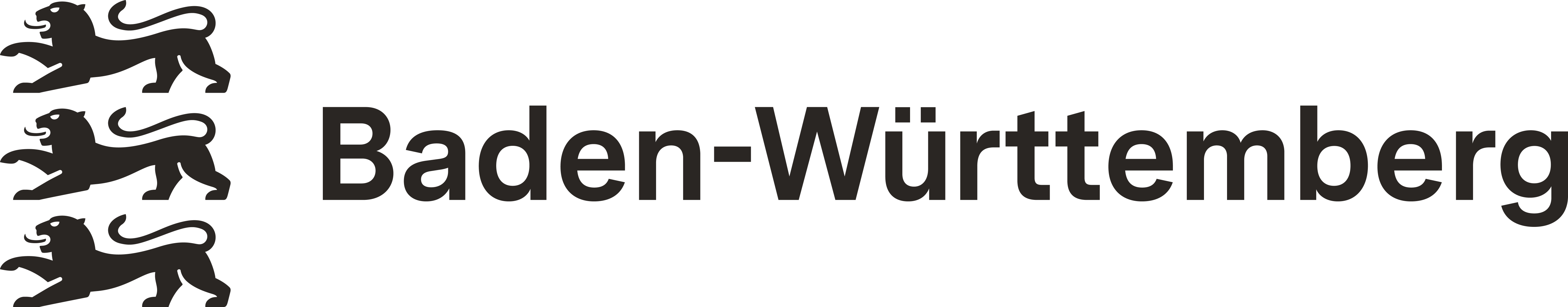}
\end{minipage}
\hfill
\begin{minipage}{0.24\textwidth}
\centering
\includegraphics[width=\textwidth]{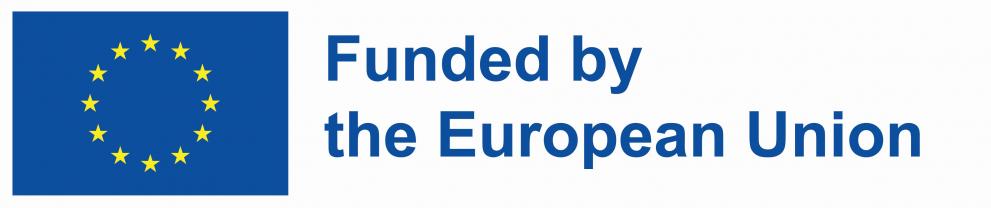}
\end{minipage}
\hfill
\begin{minipage}{0.24\textwidth}
\centering
\includegraphics[width=\textwidth]{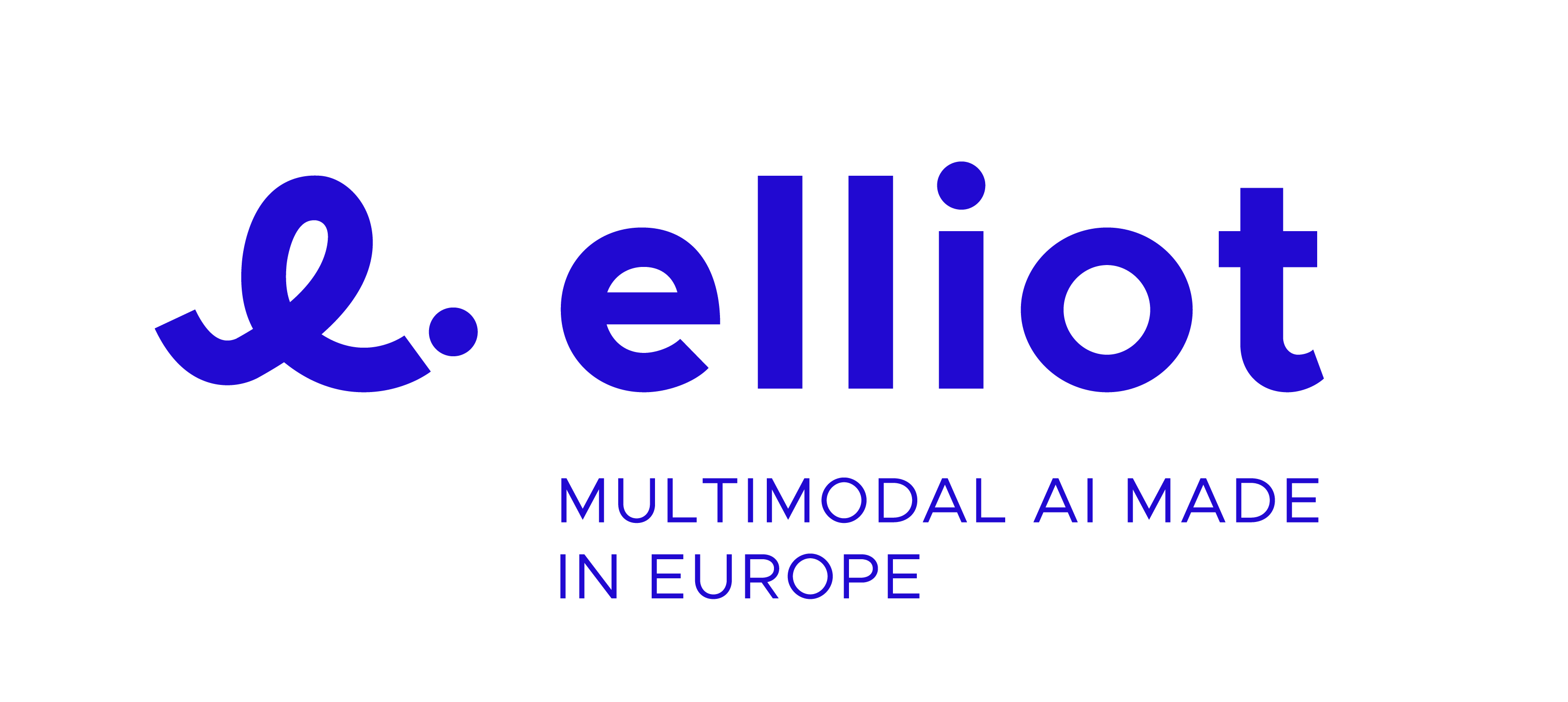}
\end{minipage}
\hfill
\begin{minipage}{0.24\textwidth}
\centering
\includegraphics[width=\textwidth]{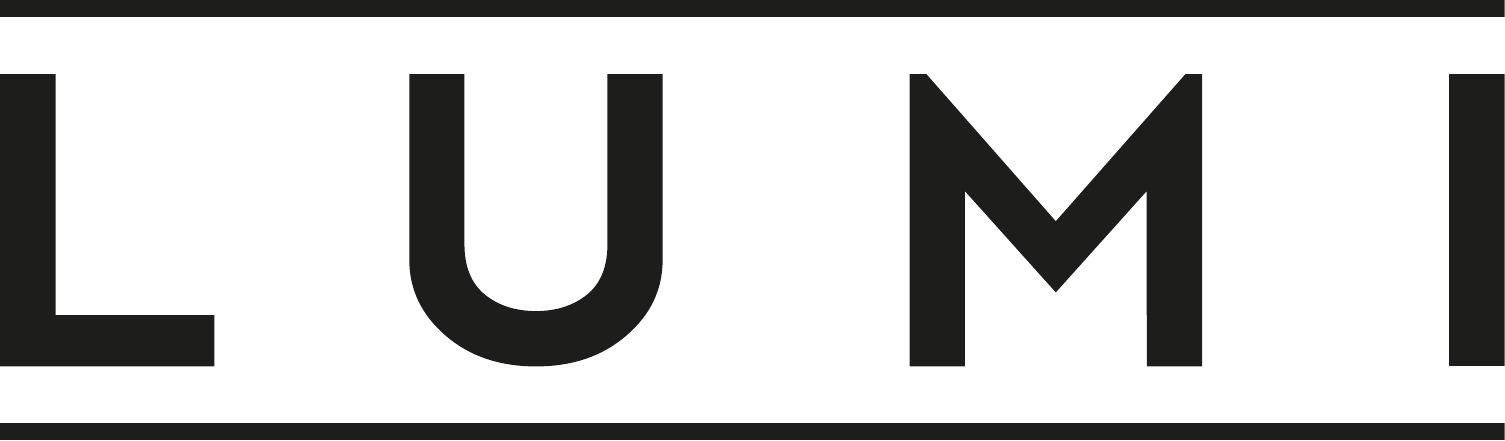}
\end{minipage}
\hfill ~
\end{center}

\newpage

\bibliographystyle{tmlr-initials}
\bibliography{bib/lib,bib/local,bib/proc,bib/shortstrings}

\newpage

\appendix

\section{Detailed Experimental Setup}
\label{app:experimental_setup}

\subsection{Training Configuration}
\label{app:training-config}

We use AdamW with $\beta_1=0.9$, $\beta_2=0.999$, $\epsilon=10^{-6}$, and gradient-norm clipping at $1.0$, optimizing cross-entropy loss.
Training uses \texttt{float32} precision without mixed precision or learning-rate warmup.
We implement effective batch sizes through gradient accumulation, choosing a micro-batch size that fits within $90\%$ of available GPU memory.
The data-gathering experiments were conducted primarily on NVIDIA RTX 2080 Ti and L40S GPUs, using one GPU per training job.
The data-gathering runs required approximately 80{,}500 training GPU-hours, excluding evaluation and the extrapolation runs.

\subsection{Search Space and Data Collection}
\label{app:search-space}

\begin{table}[t]
  \centering
  \small
  \caption{Hyperparameter values searched over all runs with a training budget up to $10^{16}$ FLOPs. Fixed in all
  runs: AdamW, MLP hidden size $4\times$ the embedding size. Parameters follow from embedding size and layers;
  max.\ cells is max.\ features $\times$ max.\ datapoints.}
  \begin{tabular}{l p{0.62\linewidth}}
    \hline
    Hyperparameter & Values \\
    \hline
    Learning rate & $\{5\cdot10^{-5}$, $1.44\cdot10^{-4}$, $4.16\cdot10^{-4}$, $1.2\cdot10^{-3}$, $2.1\cdot10^{-3}$, $3.47\cdot10^{-3}$, $5.72\cdot10^{-3}$, $7\cdot10^{-3}$, $1\cdot10^{-2}$, $3.47\cdot10^{-2}$, $7.07\cdot10^{-2}$, $5\cdot10^{-1}\}$ \\
    Weight decay & \{0, 0.0001, 0.1\} \\
    Batch size & \{32, 64, 128\} \\
    Embedding size & \{8, 16, 32, 64\} \\
    Layers & \{1, 2, 4, 6, 8, 10\} \\
    Attention heads & \{2, 4\} \\
    Max.\ features per table & \{2, 4, 8, 16\} \\
    Max.\ datapoints per table & \{100, 200, 400, 800, 1200\} \\
    \hline
    \multicolumn{2}{l}{\emph{Derived quantities (not searched directly)}} \\
    \hline
    Parameters $N$ & [1,826--553,482] (20 distinct) \\
    Max.\ cells per table & \{200, 400, 800, 1600, 3200, 6400, 9600, 12800, 19200\} \\
    \hline
  \end{tabular}
  
  \label{tab:hyperparameters}
\end{table}

Table~\ref{tab:hyperparameters} summarizes the eight hyperparameters explored in our scaling analysis.
We conducted manually designed sweeps at successive compute budgets, refining the search based on earlier results while extending the range of model and synthetic prior sizes.
In total, we collected 6{,}075 data-gathering runs across four compute budgets, as summarized in Table~\ref{tab:sweeps}.
We additionally trained 45 configurations at $10^{17}$ FLOPs, selected around the fitted scaling prescriptions and therefore not restricted to the values in Table~\ref{tab:hyperparameters}, and excluded them from scaling-law fitting.

\begin{table}[t]
    \centering
    \small
    \caption{Number of training runs by nominal compute budget. The $10^{17}$-FLOP runs are reserved for extrapolation evaluation.}
    \label{tab:sweeps}
    \begin{tabular}{lccccc}
        \toprule
        Budget (FLOPs) & $10^{13}$ & $10^{14}$ & $10^{15}$ & $10^{16}$ & $10^{17}$ \\
        \midrule
        Runs & 4{,}050 & 1{,}651 & 288 & 86 & 45 \\
        \bottomrule
    \end{tabular}
\end{table}

\paragraph{Prior Ablation}
For Figure~\ref{fig:scaling-data-prior}, we compare three subsets of training configurations.
The \emph{fixed narrow prior} and \emph{fixed wide prior} restrict configurations to maximum prior dimensions of $4$ features and $100$ rows, and $16$ features and $400$ rows, respectively.
The \emph{scaled prior} uses the unrestricted frontier, allowing prior dimensions to vary with compute.
The fixed wide prior is the largest considered in this ablation, rather than the largest configuration in the full search space.

\subsection{FLOP calculation}
\label{app:flops}

\paragraph{FLOP Accounting}
We estimate training compute using an architecture-specific count of
matrix multiplication FLOPs. Let $T$ denote the number of rows,
$d$ the number of columns (including the target), $n$ the number
of labeled context rows, $E$ the embedding dimension, and $L$
the number of layers. With MLP expansion factor $M=4$ and
$K=10$ output classes, the FLOPs for one forward and backward
pass over a table are
\begin{equation}
\begin{aligned}
f(T,n,d) = 3\Big[
    &2TdE \\
    &+ L\big((12+4M)TdE^2 + 4ndE^2
             + 4ET(d^2+nd)\big) \\
    &+ 2ME(E+K)(T-n)
\Big].
\end{aligned}
\label{eq:flops}
\end{equation}
We count $2abc$ FLOPs for multiplying matrices of dimensions
$a\times b$ and $b\times c$, and approximate the backward pass
as twice the forward-pass cost. The formula includes encoders,
attention projections and products, MLPs, and the decoder,
while excluding elementwise operations.

Since synthetic table dimensions vary during training, we compute
cumulative FLOPs by summing $B_k f(T_k,n_k,d_k)$ over all
micro-batches $k$, where $B_k$ is the number of tables in each
micro-batch, including those used for gradient accumulation.
Earlier FLOP estimates obtained with \texttt{calflops} omitted attention products; we therefore recomputed the training compute of affected runs using Equation~\ref{eq:flops}.

\paragraph{FLOP-Grid Alignment}
Evaluations are recorded after optimizer steps and therefore do not necessarily coincide with the intended compute budgets. Recomputing FLOPs further shifted previously recorded evaluations away from their original milestones. To ensure consistent comparisons, we align all runs to a common grid $C_k=10^{11+k/4}$, $k=0,\ldots,32$. For each grid point, we retain an observed evaluation if it lies within $0.01$ dex ($2.3\%$); otherwise, we interpolate linearly in $\log_{10}$ FLOPs between adjacent evaluations separated by at most $0.35$ dex. We discard unmatched grid points without extrapolation and exclude runs with fewer than two valid evaluations. The same procedure applies to evaluation metrics and cumulative training cells, ensuring consistent compute alignment.
\subsection{Evaluation Protocol}

\paragraph{Synthetic Validation Data}
We generate two held-out sets of synthetic classification datasets using the TabICLv1 prior~\citep{qu2025tabicl}. The small validation set contains $2{,}048$ datasets with $500$ to $1{,}000$ rows and $5$ to $50$ features, while the medium set contains $1{,}024$ datasets with $1{,}000$ to $3{,}000$ rows and $50$ to $150$ features. Each dataset has at most $10$ classes. Datasets are generated in batches of $32$, with the number of rows sampled uniformly within the corresponding range. For each batch, we sample an in-context training fraction $p\sim\mathcal{U}(0.1,0.9)$, assigning the first $\lfloor pN\rfloor$ rows to training and the remaining rows to testing. Both validation sets are generated once and reused across all checkpoints. We compute the mean loss within each set and average the two set-level losses equally to obtain the synthetic validation score.

\paragraph{Evaluation-Set Selection Experiment}
For each of the 21 evaluation datasets, we hold out that dataset and sample $n\in\{1,2,3,5,8,12,16,20\}$ datasets from the remaining 20, repeating the selection 20 times.
At each compute budget, we select the configuration with the best mean performance on the selection datasets and record both its selection-set performance and its performance on the held-out dataset.
We average these scores separately across the 21 held-out datasets and construct the corresponding performance frontiers by independently applying a running maximum for ROC AUC or a running minimum for NLL over compute budgets.
We define the generalization gap as the difference between the resulting frontiers, with positive values indicating higher selection-set ROC AUC or lower selection-set NLL.
We average this gap over the 20 compute budgets and report its median and 10--90\% quantiles across the 20 random repetitions.
For comparison, we also report the fitting residuals of scaling laws fitted to the selection-set frontiers.

\section{Scaling-Law Fitting}
\label{app:scaling_fits}

\paragraph{Pareto Frontier Construction.}
For each evaluation metric, we collect performance measurements
from the training configurations and their intermediate checkpoints.
For metrics to be maximized and bounded above by $1$, such as
accuracy and ROC AUC, we transform the observations using $L=1-y$.
For loss metrics such as NLL, we use the original values.

We construct the empirical Pareto frontier over training compute
and performance. An observation $(C_i,L_i)$ is dominated if another
observation $(C_j,L_j)$ satisfies $C_j\leq C_i$ and $L_j\leq L_i$,
with at least one strict inequality.
We retain only non-dominated observations for fitting, thereby
approximating the best observed performance achievable within
each compute budget.

\paragraph{Parameter Estimation.}

We fit the scaling law in Equation~\ref{eq:scaling} to the
Pareto-optimal observations.
Let $\mathcal{P}$ denote the indices of these observations and
$\theta=(A,\alpha,L_0)$ the scaling-law parameters.
Following \citet{li2025mis}, we minimize the Huber loss
in logarithmic space:
\begin{equation}
    \hat{\theta}
    =
    \arg\min_{\theta\in\Theta}
    \frac{1}{|\mathcal{P}|}
    \sum_{i\in\mathcal{P}}
    \rho_\delta\left(
        \log L_i -
        \log\left(A C_i^{-\alpha}+L_0\right)
    \right),
    \label{eq:scaling_fit_objective}
\end{equation}
where $\rho_\delta$ denotes the Huber loss with $\delta=10^{-3}$, and $\Theta=\{(A,\alpha,L_0): A>0,\ L_0>0,\ \alpha\in\mathbb{R}\}$.
We optimize the transformed parameters $a=\log A$, $\ell_0=\log L_0$, and $\alpha$, evaluating predictions using $\log L(C)=\operatorname{logsumexp}(\ell_0,a-\alpha\log C)$ for numerical stability.
We minimize the objective using L-BFGS-B without explicit box bounds and retain the lowest-loss solution across $105$ initializations.
The initializations form a Cartesian grid over $\ell_0\in\{0,0.1,0.5\}$, $a\in\{0.1,1,2,5,10,50,100\}$, and $\alpha\in\{-1,-0.5,0,0.5,1\}$.
We do not constrain the sign of $\alpha$ during optimization, and all fitted exponents are positive, consistent with $\alpha > 0$ in Equation~\ref{eq:scaling}.
The same fitting procedure is used for synthetic validation loss, real-world NLL, and transformed ROC AUC.

We additionally derive compute-dependent scaling prescriptions $h(C)=aC^{\gamma}$ for all eight search parameters and derived quantities, including model parameter count, total training cells\footnote{We treat \textit{cells} as the atomic unit of a dataset, i.e., a single cell is a $1 \times 1$ table.}, and maximum prior table size.
Following Approach 1 of \citet{hoffmann-neurips22a}, we fit these power laws by ordinary least squares in log-log space to the configurations on each metric's Pareto frontier.

\paragraph{Extrapolation Protocol}
We assess extrapolation by evaluating both scaling-law predictions and compute-optimal scaling prescriptions at unseen compute budgets.
For each metric, we fit Equation~\ref{eq:scaling} to Pareto-frontier observations up to $10^{16}$ FLOPs and evaluate predictions at four held-out budgets between $10^{16.25}$ and $10^{17}$ FLOPs.
We additionally train $45$ configurations at a nominal budget of $10^{17}$ FLOPs, selected around the fitted scaling prescriptions and excluded from fitting.
These configurations are included when constructing the observed performance frontiers at the held-out budgets.
We measure extrapolation error using the mean absolute error (MAE) between predicted and best observed performance across the four budgets.

\begin{table}[t]
    \centering
    \small
    \caption{Scaling-law fit quality and extrapolation to unseen compute budgets.
    Fit $R^2$ is computed on the training frontier in the original metric space, while extrapolation MAE is averaged over four held-out budgets from $10^{16.25}$ to $10^{17}$ FLOPs.}
    \label{tab:extrapolation}
    \begin{tabular}{lrrrr}
        \toprule
        Metric & Observed at $10^{17}$ & Predicted at $10^{17}$ & Fit $R^2$ & Extrapolation MAE \\
        \midrule
        Synthetic validation loss & 0.770 & 0.769 & 0.997 & 0.001 \\
        Real-world NLL            & 0.309 & 0.289 & 0.986 & 0.011 \\
        Real-world ROC AUC        & 0.877 & 0.873 & 0.991 & 0.004 \\
        \bottomrule
    \end{tabular}
\end{table}

Table~\ref{tab:extrapolation} shows high in-sample fitting accuracy across all three metrics, with $R^2\geq 0.986$.
Extrapolation errors remain small for synthetic validation loss and real-world ROC AUC, with MAEs of $0.001$ and $0.004$, respectively.
Real-world NLL exhibits a larger extrapolation error of $0.011$, with the fitted law underestimating NLL by $0.020$ at $10^{17}$ FLOPs.

\begin{figure}[ht]
  \centering
  \includegraphics[width=0.8\linewidth]{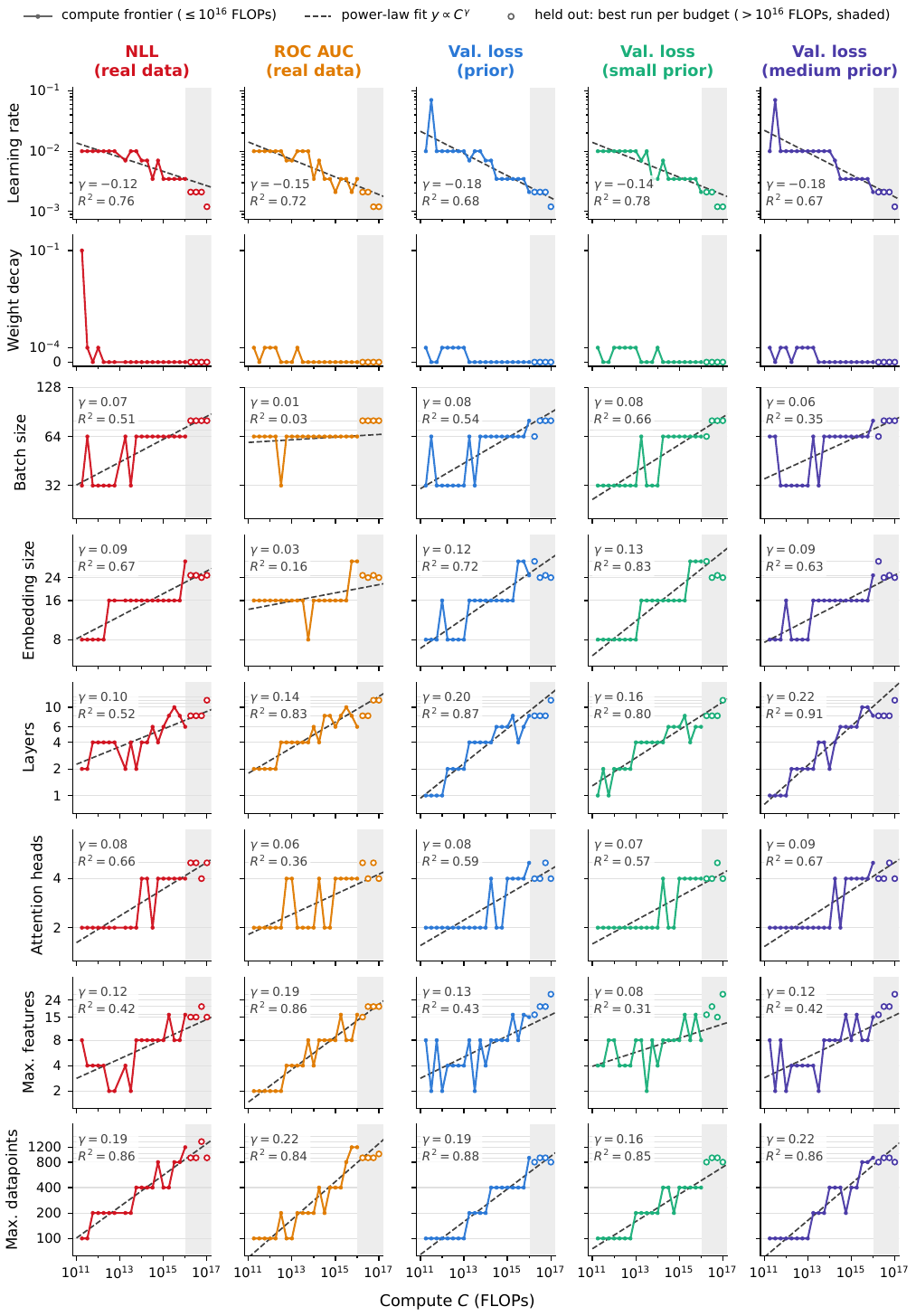}
  \caption{
  Compute-optimal values of the searched hyperparameters. 
  Each column is one metric, each row one hyperparameter. 
  A filled point is the run on the compute frontier of that metric at the given FLOP budget, i.e. the best run at that budget that also beats every smaller budget; light horizontal lines mark the values in the search space. 
  The dashed line is a power law $y = c\cdot C^{\gamma}$ fitted by least squares in log–log space to the frontier points up to $10^{16}$ FLOPs, with the exponent $\gamma$ and the in-sample $R^2$ given in each panel. 
  Open points in the shaded region are the best runs at the four held-out budgets ($1.8\cdot10^{16}$ to $10^{17}$ FLOPs), which do not enter the fit. 
  Weight decay is not fitted because the frontier contains zero.}
  \label{fig:trends_searched}
\end{figure}
\begin{figure}[ht]
  \centering
  \includegraphics[width=0.8\linewidth]{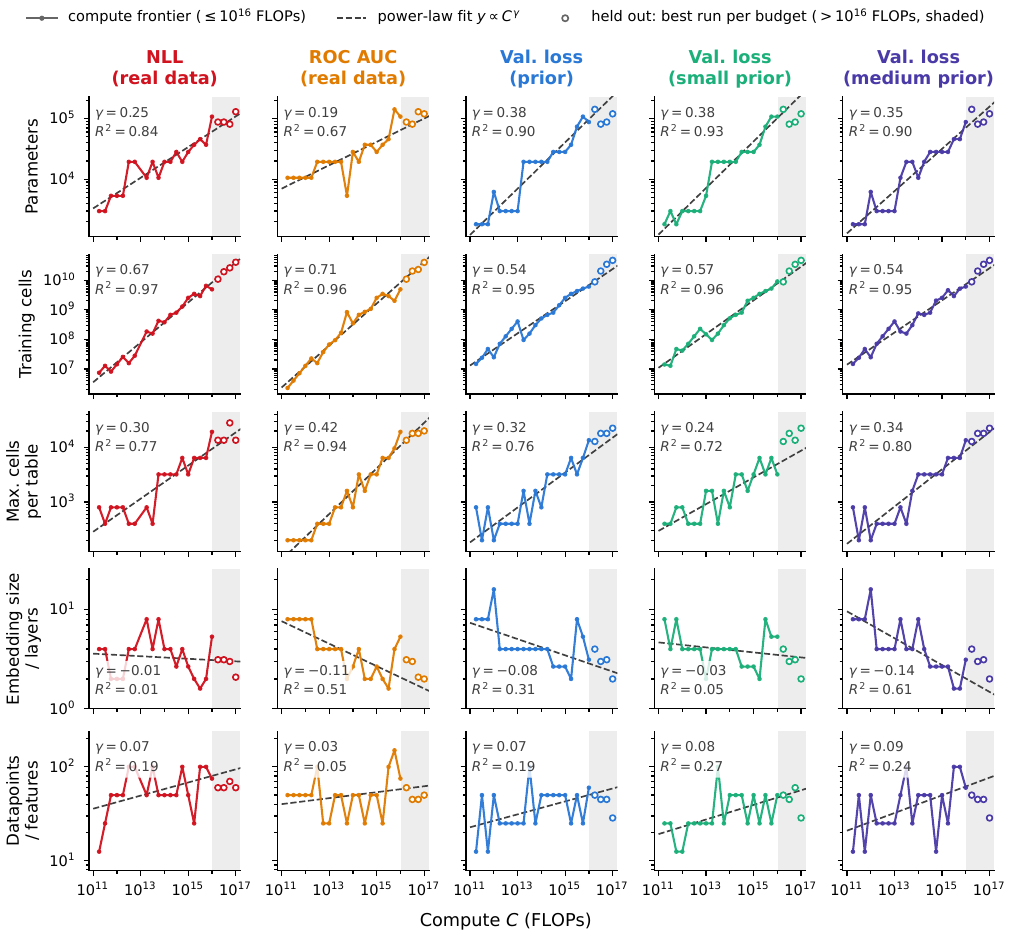}
  \caption{Compute-optimal values of quantities derived from the searched hyperparameters: parameter count, number of training cells seen, largest table in the prior (max. features $\times$ max. datapoints), and the ratios embedding size / layers and datapoints / features. 
  Markers, lines and the shaded held-out region are as in Figure~\ref{fig:trends_searched}. Training cells and parameters follow a power law closely ($R^2 \geq 0.84$ for all metrics except ROC AUC on parameters), whereas the two ratios show no consistent trend with compute.}
  \label{fig:trends_derived}
\end{figure}

\section{Scaling Law Fit for Phase Shifting ROC AUC}

\subsection{Fitting the Shifted ROC AUC Frontier}
\label{sec:big-prior-forms}

\begin{figure}[ht]
    \centering
    \includegraphics[width=0.75\linewidth]{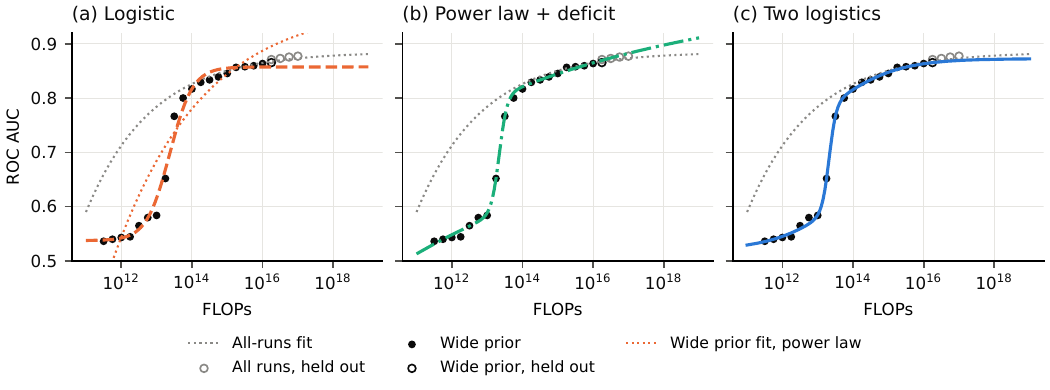}
    \caption{ROC AUC frontier for the largest synthetic prior
    and three fitted functional forms: (a) logistic,
    (b) power law with a logistic deficit, and
    (c) two logistics. Hollow markers denote held-out
    observations; the dotted gray curve shows the
    fitted frontier over all configurations.}
    \label{fig:big-prior-fits}
\end{figure}

The ROC AUC frontier for the largest synthetic prior
(16 features, 400 rows) exhibits an initial plateau,
followed by a sharp increase and slower subsequent improvement (Figure~\ref{fig:big-prior-fits}).
A single power law or logistic function cannot adequately capture
both phases. We therefore compare several functional forms,
including broken neural scaling laws~\citep{caballero-iclr23a},
and report three representative alternatives: a logistic $g_{\mathrm{log}}$,
a power law with a logistic deficit $g_{\mathrm{def}}$, and a sum of two
logistics $g_{\mathrm{2log}}$~\citep{meyer1994bi}.

Let $x=\log_{10}C$, $z=C/10^{13}$, and
$\sigma(t)=(1+e^{-t})^{-1}$. The three forms are
\begin{align}
g_{\mathrm{log}}(C)
    &= a_{\min}+(a_{\max}-a_{\min})
       \sigma(\kappa(x-x_0)),\\
g_{\mathrm{def}}(C)
    &= 1-L_0-bz^{-\alpha}
       -h\sigma(-\kappa(x-x_0)),\\
g_{\mathrm{2log}}(C)
    &= a_{\min}
       +\sum_{j=1}^{2}
       h_j\sigma(\kappa_j(x-x_j)).
\end{align}

We fit each form to the 19 frontier observations using Huber loss ($\delta=10^{-3}$) in ROC AUC space and bounded L-BFGS optimization with multiple initializations.
We assess extrapolation using seven increasing compute cutoffs and an additional held-out observation at $1.8\times10^{16}$ FLOPs. 
We also compare extrapolations against the fitted frontier over all configurations.

\begin{table}[t]
\centering
\small
\caption{Comparison of representative functional forms.
Maximum excess measures the largest difference between the fitted prior-specific curve and the all-configurations fit over $10^{11}$--$10^{19}$ FLOPs.
Negative values indicate that the prior-specific fit
remains below the reference curve.}
\begin{tabular}{lccc}
\toprule
Form & RMSE & Backtest MAE & Max.\ excess \\
\midrule
Logistic            & 0.0140 & 0.024 & $+0.008$ \\
Power law + deficit & 0.0048 & \textbf{0.004} & $+0.030$ \\
Two logistics       & \textbf{0.0038} & 0.008 & $\mathbf{-0.002}$ \\
\bottomrule
\end{tabular}
\label{tab:big-prior-results}
\end{table}

As shown in Figure~\ref{fig:big-prior-fits} and Table~\ref{tab:big-prior-results}, the logistic fails to capture both phases, while the power law with a logistic deficit achieves lower backtest error but extrapolates above the all-configurations fit.
The two-logistic formulation achieves the lowest fitting error and remains below the all-configurations fit over the examined range. 
We therefore use it for the largest-prior ablation.

\end{document}